\documentclass[final]{cvpr}

\usepackage{times}
\usepackage{epsfig}
\usepackage{graphicx}
\usepackage{amsmath}
\usepackage{amssymb}
\usepackage{tabularx}

\usepackage[pagebackref=true,breaklinks=true,colorlinks,bookmarks=false]{hyperref}

\def\cvprPaperID{****} 
\def\confYear{CVPR 2021}

\begin{document}

\title{TinyAction Challenge: Recognizing Real-world Low-resolution Activities in Videos
}

\author{
Praveen Tirupattur\textsuperscript{*}, Aayush J Rana\textsuperscript{*}, Tushar Sangam\textsuperscript{*},\\
Shruti Vyas\textsuperscript{\rm \dag}, Yogesh S Rawat\textsuperscript{\rm \ddag}, Mubarak Shah\textsuperscript{\rm \ddag}\\
\textit{Center for Research in Computer Vision}\\
\textit{University of Central Florida, Orlando, Florida, USA}\\
{\tt\small Email: \textsuperscript{*}[praveentirupattur, aayushjr, tusharsangam]@knights.ucf.edu, \textsuperscript{\rm \dag}shruti@crcv.ucf.edu}\\
{\tt\small \textsuperscript{\rm \ddag}[yogesh, mubarak.shah]@ucf.edu}
}

\maketitle

\begin{abstract}
   This paper summarizes the TinyAction challenge \footnote{\url{https://www.crcv.ucf.edu/tiny-actions-challenge-cvpr2021}} which was organized in ActivityNet workshop at CVPR 2021. This challenge focuses on recognizing real-world low-resolution activities present in videos. Action recognition task is currently focused around classifying the actions from high-quality videos where the actors and the action is clearly visible. While various approaches have been shown effective for recognition task in recent works, they often do not deal with videos of lower resolution where the action is happening in a tiny region. However, many real world security videos often have the actual action captured in a small resolution, making action recognition in a tiny region a challenging task. In this work, we propose a benchmark dataset, TinyVIRAT-v2 \footnote{\url{https://www.crcv.ucf.edu/tiny-actions-challenge-cvpr2021/data/TinyVIRAT-v2.zip}}, which is comprised of naturally occuring low-resolution actions. This is an extension of the TinyVIRAT dataset \cite{demir2021tinyvirat} and consists of actions with multiple labels. The videos are extracted from security videos which makes them realistic and more challenging. We use current state-of-the-art action recognition methods on the dataset as a benchmark, and propose the TinyAction Challenge.
\end{abstract}


%

\section{Introduction}

In recent years, action recognition from videos has become widely applied in security analysis security and automation tasks. The availability of large-scale datasets and the progress of neural networks have provided significant improvement to video action recognition task. Datasets with multiple actors and actions such as UCF-101 \cite{soomro2012ucf101}, Kinetics \cite{smaira2020shortkinetics700, kay2017kinetics}, AVA \cite{gu2017ava}, YouTube-8M \cite{abu2016youtube} and Moments-in-time \cite{monfortmoments} provide a large set of data with higher versatility for training neural networks. This has enabled several state-of-the-art architectures such as C3D \cite{tran2015c3dfb}, I3D \cite{zisserman2017quoi3d}, ResNet-3D \cite{hara2018resnet3d} and R2+1D \cite{tran2018closer} which have been effective at recognizing the correct actions. While development of such architectures and larger datasets have improved action recognition in videos, it is ignoring a large portion of real life videos where the actions are occurring at a distance with a smaller resolution. The existing research in action recognition is mostly focused on high-quality videos where the action is distinctly visible. Recognizing such actions is a challenging problem since the available architectures are not designed to handle low-resolution regions with less information. Due to a lack of appropriate architectures and datasets that focus on such low-resolution actions, their performance is still far from satisfactory when the action is not distinctly visible.

\begin{figure*}
\begin{center}
\label{fig:teaser}
\includegraphics[width=0.95\linewidth]{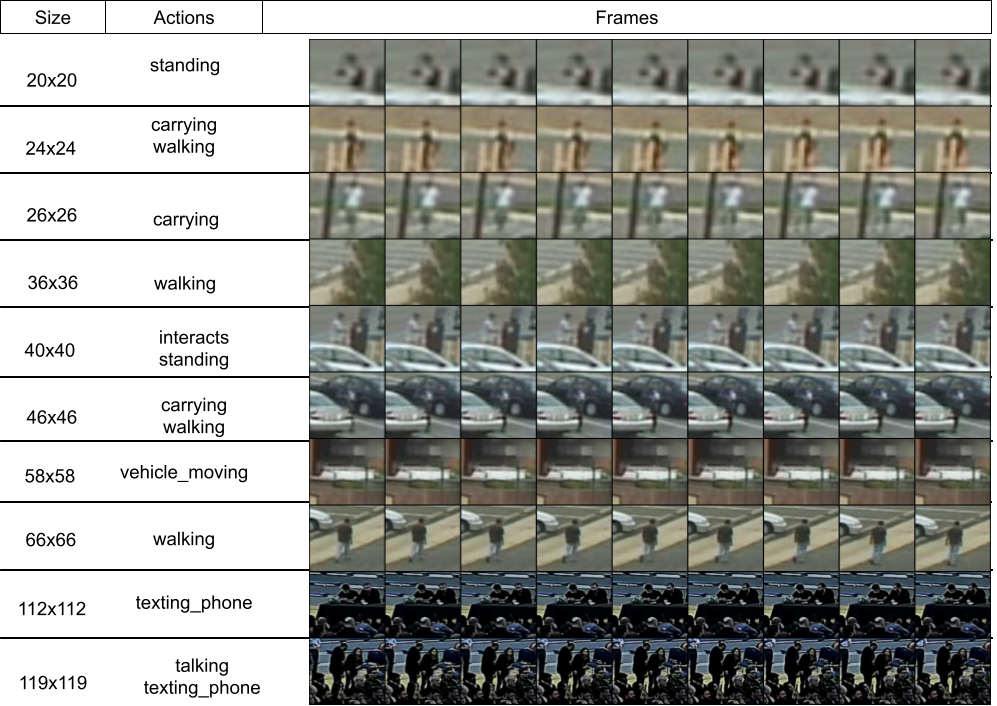}
\end{center}
  \caption{Sample video frames for various actions from \textit{TinyVIRAT-v2} dataset. \textit{TinyVIRAT-v2} is a multi-class multi-label dataset with multiple actions occurring simultaneously in a single video. 
  }
\end{figure*}


\begin{table*}[ht!]
    \centering
    \caption{Dataset statistics. ANF: Average number of frames, ML: Multi-label, NC: Number of classes, and NV: Number of Videos.}
    \begin{tabularx}{\linewidth}{c | c | c | c | c | c | c | c | c}
    \hline
    Dataset & Resolution & ANF & ML & NC & NV & Train & Val & Test \\
    \hline
    \hline 
    UCF-101 \cite{soomro2012ucf101}   & 320x240           & 186.50   & No  & 101 & 13320 & 9537 & - & 3783 \\
    HMDB-51 \cite{kuehne2011hmdb}    & 320x240           & 94.49    & No  & 51  & 7000 & 3570 & 1530 & - \\ 
    AVA \cite{gu2017ava} & 264x440 - 360x640 & 127081.66 & Yes & 80 & 385,446(272) & 210,634 & 57,371 & 117,441 \\
    \hline
    \textbf{TinyVIRAT \cite{demir2021tinyvirat}}  & 10x10 - 128x128   & 93.93    & Yes & 26  & 12829 & 7663 & - & 5166  \\
    \hline
    \textbf{TinyVIRAT-v2}  & 10x10 - 128x128   & 76.14    & Yes & 26  & 26355 & 16950 & 3308 & 6097  \\   
    \hline
    \end{tabularx}
    
    \label{table:dataset_vs_stats}
\end{table*}

The ActivityNet challenge has seen a wide range of tasks relevant to action recognition, ranging from temporal activity recognition to spatio-temporal action detection. However, in all the tasks we have seen so far, the focus has rarely been on low-resolution activities. In real-world security environments, the actions in videos are captured at a wide range of resolutions, where most activities occur at a distance at a small resolution. Contrary to this, most widely used datasets contain high-resolution videos where the occurring activities cover most of the region.

In this work, the focus is on recognizing tiny actions in low-resolution videos. The existing approaches addressing this issue perform their experiments on artificially created datasets where the high-resolution videos are down-scaled to a smaller resolution to create a low-resolution sample. However, re-scaling a high-resolution video to a lower- resolution does not reflect real world low-resolution video quality. Real world low-resolution videos suffer from grain, camera sensor noise, and other factors, which are not not present in the down-scaled videos. 

We address this problem via a two-pronged approach. Firstly, we provide the TinyVIRAT-v2 dataset, a benchmark dataset for activity recognition which contains natural low-resolution activities. Then we host the TinyAction Challenge to create a competitive opportunity for the research community to focus on low-resolution action recognition task and develop specific architectures to tackle its various challenges. 

The TinyVIRAT-v2 dataset is built upon the existing TinyVIRAT dataset \cite{demir2021tinyvirat} and consists of realistic low-resolution action videos extracted from security videos of VIRAT \cite{oh2011large} and MEVA \cite{Corona_2021_WACV} dataset. This is a multi-label dataset with multiple actions per video clip which makes it even more challenging. In addition to TinyVIRAT, it also includes indoor scenes making this problem more challenging and realistic.

\begin{figure*}
\begin{center}
\includegraphics[width=0.95\linewidth]{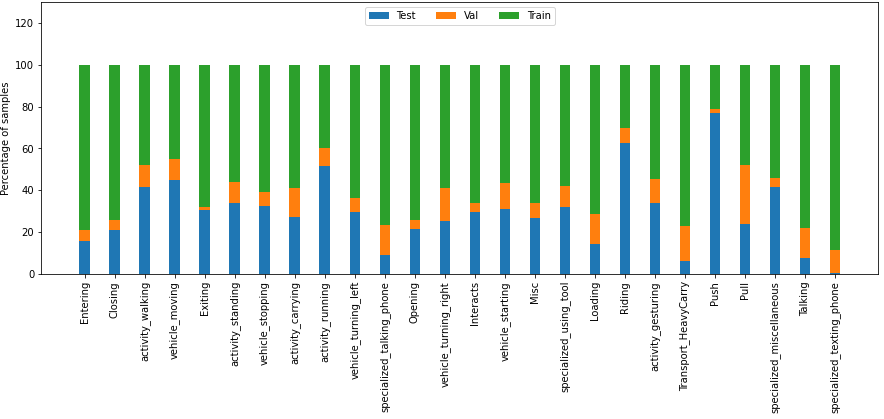}
\end{center}
  \caption{Number of samples per action class across the train, validation \& test split.}
  \label{fig:class_wise_distribution}
\end{figure*}

\begin{figure*}
\begin{center}
\includegraphics[width=0.95\linewidth]{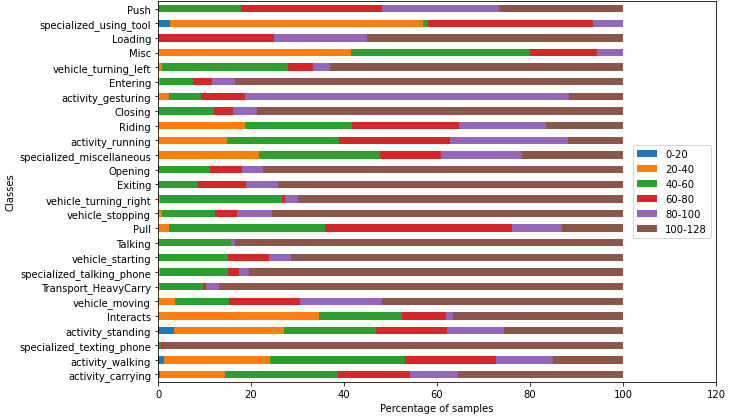}
\end{center}
  \caption{Class wise sample distribution by resolution. We group the samples into six groups based on their resolution (0-20, 20-40, 40-60, 60-80, 80-100, 100-128) for each class.}
  \label{fig:resolution_wise_distribution}
\end{figure*}

\begin{figure*}
\begin{center}
\includegraphics[width=\linewidth]{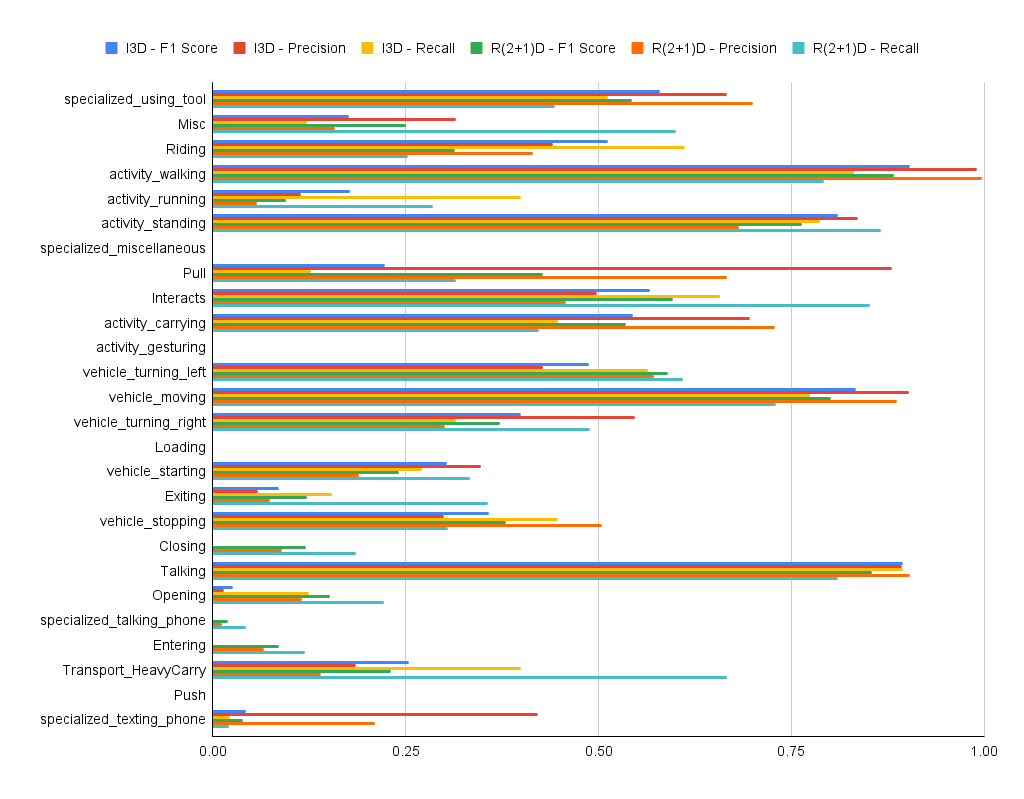}
\end{center}
  \caption{Per-class performance of the top-2 baseline models.}
  \label{fig:perclass_performance}
\end{figure*}

\begin{figure*}
\begin{center}
\includegraphics[width=0.9\linewidth]{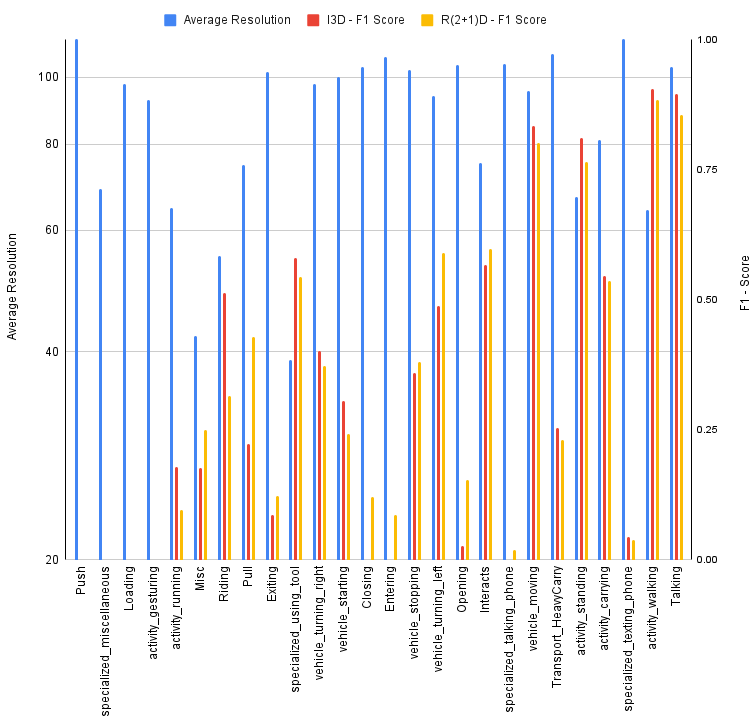}
\end{center}
  \caption{Comparison of performance of the baseline models with the average resolution for each action class.}
  \label{fig:resolution_vs_performance}
\end{figure*}

\begin{figure*}
\begin{center}
\includegraphics[width=0.9\linewidth]{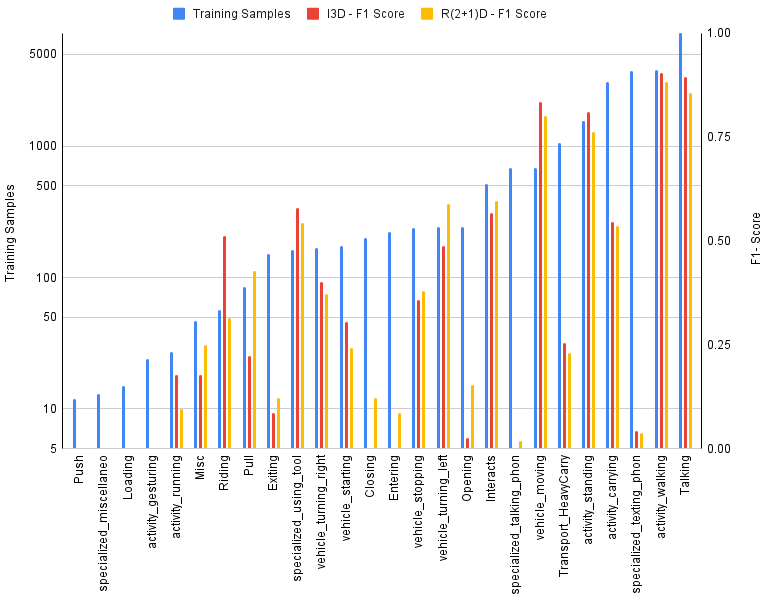}
\end{center}
  \caption{Comparison of performance of the baseline models with the number of training samples for each action class.}
  \label{fig:count_vs_performance}
\end{figure*}

\section{TinyAction Challenge}

This challenge is a first of its kind for low-resolution action recognition task. Our goal is to generate interest in the research community for such action recognition task which often is overlooked in other large-scale action datasets. As modern security and analysis videos often have multiple actions occurring in a low-resolution region further away from the camera, it is essential to bridge the gap between good action recognition architectures and real-world videos with low-resolution actions. 

\subsection{TinyVIRAT Dataset}

Most of the existing action recognition datasets contain high
resolution, actor centric videos \cite{7298839}, \cite{6909619}, \cite{kay2017kinetics}, \cite{sigurdsson2016hollywood}, \cite{abuelhaija2016youtube8m}, \cite{6247801},
\cite{gu2017ava}, \cite{7298698}, \cite{Murthy_2013_ICCV_Workshops}, \cite{damen2018scaling}. For example, Kinetics \cite{kay2017kinetics}, Charades \cite{sigurdsson2016hollywood},
Youtube-8M \cite{abuelhaija2016youtube8m} are collected from Youtube videos where
actions cover most of the image regions in every frame of a
video. Using these videos to create low-resolution benchmark
datasets does not reflect real world situation, and it is not
appropriate as they generally contain larger actors.
In the real world, we encounter low-quality actions mostly
in security video clips where the camera placed in a
distant place. Even though security camera is capable
of recording high-quality video, if an action happens far
away from the camera, it will suffer from lack of details.
Thus, security videos are the perfect candidate for this
problem. VIRAT dataset has naturally
occurring tiny actors which is well suited for low-resolution
action recognition task.

We introduce TinyVIRAT dataset which is based on VIRAT \cite{oh2011large} dataset for real-life tiny action recognition problems.
VIRAT dataseta is a natural candidate for low-resolution actions
but it contains a large variety of different actor sizes and it is a
very complex since actions can happen any time in any spatial
position. To focus only on low-resolution action recognition
problem, we crop small action clips from VIRAT videos.
In VIRAT dataset actors can perform multiple actions and
temporally actions can start and end at different times. Before
deciding which actions are tiny, we merged spatio-temporally
overlapping actions and created multi-label action clips. We
split these clips if the labels are changing temporally. This
steps makes sure that created clips are trimmed. We selected
clips that are spatially smaller than 128x128. Finally, long
videos are split into smaller chunks and actions which do not
have enough samples are removed from the dataset. TinyVIRAT has 7,663 training and 5,166 testing videos with 26 action labels. Table \ref{table:dataset_vs_stats} shows statistics from TinyVIRAT and several other datasets. 

\subsection{TinyVIRAT-v2 Dataset}
TinyVIRAT-v2 is an extension to TinyVIRAT dataset where we use MEVA dataset \cite{Corona_2021_WACV} to extract tiny actions. Much like TinyVIRAT, TinyVIRAT-v2 is based on security videos. We use the same strategy as we used for VIRAT to extract tiny actions from MEVA dataset. TinyVIRAT was restricted to only outdoor videos and in TinyVIRAT-v2 we also have indoor scenes which makes this problem more challenging. 
It adds new sub domain and challenge in the data, models trained on TinyVIRAT-v2 can be more generalized and can be trusted to work both on indoor and outdoor data along with the actions mentioned.

TinyVIRAT-v2 has 16950 videos in train, 3308 videos in validation and 6097 videos in test split.
Table \ref{table:dataset_vs_stats} shows statistics from TinyVIRAT and several other datasets.
Fig \ref{fig:class_wise_distribution}. Number of samples per action labels and Fig \ref{fig:resolution_wise_distribution} shows resolution wise sample distribution. Samples are indicated by percentage per class.

\section{Results}


\subsection{Evaluation metrics}

TinyVIRAT-v2 has multiple labels in each sample, the submissions have to predict multiple action classes for each sample. The contestants choose a prediction threshold of their choice and only submit the occurring activities for each sample as a multi-hot vector. The submissions are evaluated using precision, recall, and F1-score. The challenge winners are determined based on the F1-score averaged over each class.


\subsection{Baseline scores}
We apply the recent state-of-the-art architectures on video action recognition on TinyVIRAT-v2 dataset and evaluate them based on their precision, recall and F1-scores. We use the base versions of I3D \cite{zisserman2017quoi3d}, Resnet-3D \cite{hara2018resnet3d}, R(2+1)D \cite{tran2018closer} and WideResNet-3D \cite{zagoruyko2016wide} and modify them to take in the low resolution input videos by removing certain pooling layers. This maintains the output feature matrix size for the final classification task. The evaluation results are shown in table \ref{table:baseline_scores}. We observe that the R(2+1)D architecture gives overall best F1-score of 0.32.

We present per-class performance of the best two baseline models, I3D and R(2+1)D, in Figure \ref{fig:perclass_performance}. In Figure \ref{fig:resolution_vs_performance}, we compare the performance of these models with the average resolution of the training samples for each class. Finally in Figure \ref{fig:count_vs_performance}, we compare the per-class performance with the total number of training samples for each class. 

\begin{table}
\begin{center}
\caption{Baseline scores using various state-of-the-art methods on \textit{TinyVIRAT-v2} dataset. The overall F1-score, precision and recall is reported for each method.}
\begin{tabularx}{\linewidth}{l | c | c | c}
\hline
Method & F1-Score & Precision & Recall\\
\hline
\hline 
I3D \cite{zisserman2017quoi3d} & 0.31 & 0.36 & 0.32  \\
Resnet-3D \cite{hara2018resnet3d} & 0.25 & 0.24 & 0.36  \\ 
R(2+1)D \cite{tran2018closer} & 0.32 & 0.34 & 0.37  \\
WideResNet-3D \cite{zagoruyko2016wide} & 0.29  & 0.29 & 0.33 \\
\hline
\end{tabularx}

\label{table:baseline_scores}
\end{center}
\end{table}

\subsection{Challenge winner scores}
The submissions for each team was evaluated using the same metric in an evaluation server. At the end of the evaluation the teams were ranked based on the overall F1-score. The top 3 team scores are shown in table \ref{table:winner_scores}.

\begin{table}[t!]
\begin{center}
\caption{Scores of the top-3 team in the \textit{TinyAction Challenge}. They are ranked based on the overall F1-score.}
\begin{tabularx}{\linewidth}{l | l | c | c | c}
\hline
\# & Team Name & F1-Score & Precision & Recall\\
\hline
\hline 
1 & DeepBlue \cite{hedelving_deepblue} & 0.47 & 0.51 & 0.49 \\
2 & ALONG \cite{along} & 0.44 & 0.50 & 0.42 \\
3 & SUST\&HKU \cite{wang12sustech} & 0.41 & 0.34 & 0.37 \\
\hline
\end{tabularx}
\end{center}

\label{table:winner_scores}
\end{table}


\section{Conclusion}
We introduce an improved low-resolution tiny action recognition benchmark dataset \textit{TinyVIRAT-v2} with natural low-resolution videos. We also organize a challenge focused on tiny action recognition named \textit{TinyAction Challenge} which allows researchers and enthusiasts to develop novel architectures aimed at improving action recognition from natural low-resolution videos. We showed that existing state-of-the-art methods do not perform well on tiny action setting as they are only trained with datasets that focus on larger action-to-frame ratios. Since these datasets exclude real-life security videos with naturally occurring low-resolution actions, \textit{TinyVIRAT-v2} provides a unique opportunity to the research community to improve methods for tiny actions. The top 3 performers of the challenge were able to significantly improve the classification scores across different metrics (F1-score, precision, recall). This challenge demonstrates the need for specific methods to improve tiny action recognition. 


\section{Acknowledgement}
This research is based upon work supported by the Office of the Director of National
Intelligence (ODNI), Intelligence Advanced Research Projects Activity (IARPA), via IARPA R\&D Contract No. D17PC00345. The views and conclusions contained herein are those
of the authors and should not be interpreted as necessarily representing the official
policies or endorsements, either expressed or implied, of the ODNI, IARPA, or the U.S.
Government. The U.S. Government is authorized to reproduce and distribute reprints
for Governmental purposes notwithstanding any copyright annotation thereon.


{\small
\bibliographystyle{ieee_fullname}
\bibliography{egbib}
}

\end{document}